\documentclass[a4paper,fleqn]{cas-dc}

\usepackage[authoryear,longnamesfirst]{natbib}
\usepackage[ruled]{algorithm2e}
\def\tsc#1{\csdef{#1}{\textsc{\lowercase{#1}}\xspace}}
\tsc{WGM}
\tsc{QE}
\tsc{EP}
\tsc{PMS}
\tsc{BEC}
\tsc{DE}
\begin{document}
\let\WriteBookmarks\relax
\def\floatpagepagefraction{1}
\def\textpagefraction{.001}

\shorttitle{Leveraging social media news}

\shortauthors{Zheyan Jin et~al.}

\title [mode = title]{Let Segment Anything Help Image Dehaze}                      
\tnotemark[1,2]

\tnotetext[1]{This document is the results of the research
   project funded by the National Science Foundation.}

\tnotetext[2]{The second title footnote which is a longer text matter
   to fill through the whole text width and overflow into
   another line in the footnotes area of the first page.}

\author[1]{Zheyan Jin}[type=editor,
                        auid=000,bioid=1, 
                        orcid=0000-0001-8466-7520]

\cormark[1]

\fnmark[1]

\ead{11930051@zju.edu.cn}

%\ead[url]{www.cvr.cc, cvr@sayahna.org}

%\credit{Conceptualization of this study, Methodology, Software}

\affiliation[1]{organization={Zhejiang University},
    addressline={West Lake District 38 Zhejiang University Road}, 
    city={HangZhou},
    % citysep={}, % Uncomment if no comma needed between city and postcode
    postcode={315000}, 
    country={China}}

\author[1]{Shiqi Chen}[style=chinese] 

\author[1]{Yueting Chen}[%
   role=Co-ordinator,
   ]
\fnmark[2]
\ead{chenyt@zju.edu.cn}
\credit{Data curation, Writing - Original draft preparation}

\author[1]{Zhihai Xu}
\ead{xuzh@zju.edu.cn}
% 5 author
\author[1]{Huajun Feng}
\ead{fenghj@zju.edu.cn}

\cortext[cor1]{Corresponding author}
\cortext[cor2]{Principal corresponding author}

\fntext[fn1]{This is the first author footnote. but is common to third
  author as well.}
\fntext[fn2]{Another author footnote, this is a very long footnote and
  it should be a really long footnote. But this footnote is not yet
  sufficiently long enough to make two lines of footnote text.}

\nonumnote{This note has no numbers. In this work we demonstrate $a_b$
  the formation Y\_1 of a new type of polariton on the interface
  between a cuprous oxide slab and a polystyrene micro-sphere placed
  on the slab.
  }

% 计算机语言和视觉的大模型快速发展，通过大量的数据和模型尺寸带来的性能提升给人印象深刻。但低层计算机视觉任务，如图像去雾去模糊等，从计算成本和数据上考虑还依赖于少量数据集和小尺寸transformer模型。为此我们提出了一种将大模型先验知识融入到计算机底层视觉任务的方法。我们认为雾霾退化和纹理相关，而检测大模型也和纹理相关。我们通过实验室雾霾标定实验证明了大模型能在一定条雾霾浓度下感知的纹理分布。因此，我们提出了检测灰度编码，网络通道扩充和循环网络结构来将大模型的先验融入到任何计算机视觉低层去雾网络当中。并通过不同数据集，不同算法的对比实验证明了大模型指导底层视觉任务的有效性和适用性。最后我们通过消融实验证明了灰度编码，网络通道扩充和循环网络结构的作用。在不需要额外数据和训练资源的条件下，成功通过大模型的先验知识将在底层视觉任务性能提升。
\begin{abstract}
The large language model and high-level vision model have achieved impressive performance improvements with large datasets and model sizes. However, low-level computer vision tasks, such as image dehaze and blur removal, still rely on a small number of datasets and small-sized models, which generally leads to overfitting and local optima. Therefore, we propose a framework to integrate large-model prior into low-level computer vision tasks. Just as with the task of image segmentation, the degradation of haze is also texture-related. So we propose to detect gray-scale coding, network channel expansion, and pre-dehaze structures to integrate large-model prior knowledge into any low-level dehazing network. We demonstrate the effectiveness and applicability of large models in guiding low-level visual tasks through different datasets and algorithms comparison experiments. Finally, we demonstrate the effect of grayscale coding, network channel expansion, and recurrent network structures through ablation experiments. Under the conditions where additional data and training resources are not required, we successfully prove that the integration of large-model prior knowledge will improve the dehaze performance and save training time for low-level visual tasks.

\end{abstract}

% \includegraphics{figs/grabs.pdf}

\begin{highlights}
% 发现并通过实验证明了分割大模型涌现的抗雾能力。这些能力不是数据集和训练自带的，而是通过大尺度数据集和大尺寸模型意外获得的。
\item We discovered and demonstrated the emergence of anti-fog capabilities in large-scale image segmentation models, which were not innately present in the dataset or training process but are achieved through large-scale datasets and large-scale models.
% 将大数据大模型的优势应用到小数据小模型最底层视觉去雾任务当中。通过灰度编码和通道扩增的方式，让大模型的优势可以被传递到去雾小网络当中。大模型可以帮助去雾小网络更快的训练并提升去雾结果。
\item Through grayscale coding and channel expansion, we propose a new framework for transferring the advantages of the large model to the low-level visual dehaze task engaging with small-scale data and small models, which accelerate the adaptation of specific dehaze results.
% 通过大模型和小模型讨论了不同雾霾的场景，不同雾霾数据集的难度。比较了不同雾霾场景，不同大模型的尺寸，不同小模型的尺寸对于最终去雾结果的影响。
\item We carry out a comprehensive experiment for evaluating the proposed method and comparing the impact of different model sizes on the final dehaze results under different fog scenarios.
\end{highlights}

% Keywords

\maketitle
%(a) Segment model:将雾霾图像输入大尺寸的分割模型，将其分割结果通过灰度编码的方式输出。利用大模型的涌现能力，其也能较好处理不曾见过的雾霾图像。(b) Size comparison:无论是网络尺寸和数据集大小，现有lowlevel最大的去雾模型和分割大模型都差好多个数量级。因此大模型涌现透雾能力和去雾模型的图像域转移能力可以互相帮助。(c)Dehaze: 将灰度编码的分割mask通过mask encoder 的形式加入图像去雾网络的encoder部分。
\begin{figure*}
        \centerline{\includegraphics[width=1\textwidth]{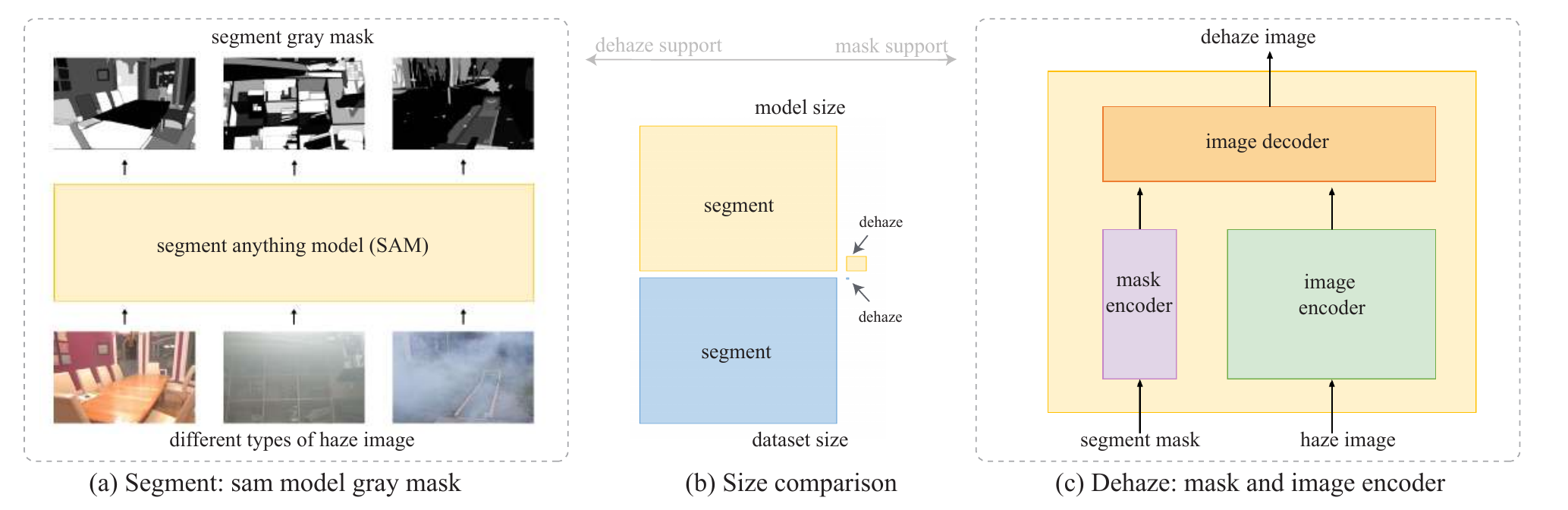}}
	\caption{\textbf{Main pipeline. (a) Segment model:} Put haze image input into a large-scale segmentation model, and output the segmentation result in grayscale encoding. By utilizing the emergence capability of large models, it can also handle haze images that have not been trained before. \textbf{(b) Size comparison:} Both in terms of network size and dataset size, the largest existing dehazing model and segmentation large model are several orders of magnitude apart. Therefore, the emergence transparency ability of large models and the image domain translation ability of the dehazing model can help each other.  \textbf{(c) Dehaze:} grayscale encoded segmentation mask is added to the encoder part of the image dehazing network through the encoder-decoder structure.}
    \label{p1pipe}
\end{figure*}

\section{Introduction}
%图像去雾是重要的计算机视觉课题之一，通过一定的图像恢复算法将成像当中的雾霾干扰去除，从而更好的进行后续计算。随着去雾算法的研究深入，浓雾，非均匀雾霾，复杂照明条件等复杂场景的出现，小尺寸模型很难处理很好的处理去雾问题。主流的去雾算法进步基本基于参数量提升的transformer模型\cite{dehazetransformer}和图像域适应\cite{domain}的方法。这些算法往往需要很多数据并训练的模型。但去雾模型往往很缺可靠实拍数据集，同时各种各样去雾任务也很难分别取训练不同的大模型。
Image dehazing is one of the important computer vision tasks, which removes the haze interference by using image restoration algorithms, allowing for better subsequent calculations. As the research on dehazing algorithms continues to advance, such as the appearance of complex scenes with thick haze, non-uniform haze, and complex lighting conditions, which small-sized models are difficult to handle well. The mainstream progress in dehazing algorithms is mainly based on transformer models with enhanced parameter counts \cite{dehazeformer} and image-domain adaptation methods \cite{domain} respectively. These algorithms often require a large number of datasets. However, dehazing models often lack reliable real-world data sets, and it is also difficult to separately train different large models for different dehazing tasks.

%随着大模型的尺寸不断发展，语言和分割大模型网络涌现除了很多超过数据本身的能力。这些涌现的能力是数据量和网络尺寸共同提升的结果。但是基于特定任务开发的计算机底层视觉的去雾模型来说，大量数据和大尺度的模型训练很难做到，从而很难分享这一网络计算红利。因此我们希望基于小模型小数据的图像去雾网络能通过大模型和大数据的图像分割模型来提升性能。让大模型来提高图像去雾的能力。
With the continuous development of large-scale models, the emergence of many abilities beyond data itself in large language and large segmentation models. These abilities have been achieved through the joint improvement of data quantity and network scale. However, image dehazing can not achieve significant improvements through large-scale and high-quality model training. It is difficult for the low-level vision dehazing model to benefit from the large model with large data size and continuous rapid development. Therefore, we hope that image dehazing networks based on small models and small data can be improved through the large image segmentation models. This allows large models to enhance the ability of image dehazing.

%我们发现了分割大模型在图像域上涌现的自适应能力，即便训练数据集当中没有针对雾霾训练的图像，通过参数量的提升，大模型可以快速弥补雾霾对于分割的性能影响。因此，对于各种雾霾场景我们使用了不同参数量的分割大模型来指导编码图像去雾模型。并通过灰度编码和通道扩充的方法让去雾小网络学会大网络的去雾能力。上述的内容使得大模型强大的泛化去雾能力应用在小尺寸的去雾网络成为可能。 
We have discovered an emergent self-adaptive ability of large-scale image segmentation models in the image domain. Even if the training dataset does not contain images specifically for fog, by increasing the network parameter, the large model can quickly compensate for the performance impact of fog on segmentation. Therefore, for various fog scenarios, we use different parameter-count large-scale segmentation models to guide the encoding image dehazing model. And through the method of grayscale coding and channel expansion, the small dehazing network can learn the dehazing ability of the large segmentation model. The above content enables the application of powerful generalization anti-haze capabilities of large models in small dehaze networks.

%我们主要的创新和贡献如下：
Our main innovations and contributions are as follows:
\begin{itemize} 
% 发现并通过实验证明了分割大模型涌现的抗雾能力。这些能力不是数据集和训练自带的，而是通过大尺度数据集和大尺寸模型意外获得的。
\item We discovered and demonstrated the emergence of anti-fog capabilities in large-scale image segmentation models, which were not innately present in the dataset or training process but are achieved through large-scale datasets and large-scale models.
% 将大数据大模型的优势应用到小数据小模型最底层视觉去雾任务当中。通过灰度编码和通道扩增的方式，让大模型的优势可以被传递到去雾小网络当中。大模型可以帮助去雾小网络更快的训练并提升去雾结果。
\item Through grayscale coding and channel expansion, we propose a new framework for transferring the advantages of the large model to the low-level visual dehaze task engaging with small-scale data and small models, which accelerate the adaptation of specific dehaze results.
% 通过大模型和小模型讨论了不同雾霾的场景，不同雾霾数据集的难度。比较了不同雾霾场景，不同大模型的尺寸，不同小模型的尺寸对于最终去雾结果的影响。
\item We carry out a comprehensive experiment for evaluating the proposed method and comparing the impact of different model sizes on the final dehaze results under different fog scenarios.
\end{itemize}

\section{Related Works}

\subsection{Image dehazing}

%图像去雾是一种低级计算视觉图像恢复。通过一些图像处理的方法消除雾霾散射对图像的负面影响。\cite{2} 使用随机森林回归器来估计雾霾。 \cite{1}最先使用暗通道先验来去除雾霾。基于深度学习的去雾方法可以分为计算中间参数和直接端到端训练两大类。计算中间参数方法，网络会推理得到中间参数，然后代入大气退化模型计算最终的无雾图像。后来的模型倾向于直接端到端学习有雾图像到无雾图像的映射。 \cite{3} 等人介绍了一种名为 DehazeNet 的端到端 CNN 网络。该模型的输入雾霾图像，输出是整个图像的透射率图，然后将透射率图和估计的全局大气光代入退化模型。 \cite{4} 等人提出了一种多尺度深度神经网络来估计透射率。 \cite{5}等人提出了阈值融合子网络，利用GAN实现图像去雾。\cite{dehazeformer}将更大参数量的transformer结构用到了去雾领域获得了更好的效果。\cite{c2pnet} 将同数据集不同雾霾图像的域变化规律用在了图像去雾上。

Image dehazing is a low-level computer vision task. It aims to eliminate the negative effects of haze scattering on images. \cite{2} used a random forest regression model to estimate the amount of haze. \cite{1} was the first to use a dark channel prior to removing haze. Deep learning-based dehazing methods can be divided into two main categories: calculation of intermediate parameters and direct end-to-end training. The former estimates intermediate parameters and then inputs them into an atmospheric degradation model to calculate the final clean image. Later models tend to learn the mapping from hazy to clean images directly. \cite{3} presented a full convolution neural network (CNN) called DehazeNet for image dehazing. The model accepts hazy images as input and produces a transmission map as its output. \cite{4} proposed a multi-scale deep neural network to estimate the transmission value. \cite{5} proposed a threshold fusion network that utilizes a generative adversarial network (GAN) for image dehazing, solving the common unreal unreality issue. \cite{dehazeformer} applied a larger parameter-count transformer structure to the dehazing field and achieved better results. \cite{c2pnet} used the domain changes between different hazy images from the same dataset for image dehazing.

%与普通图像去雾相比，浓雾，非均匀雾霾，夜间图像去雾的场景条件更为复杂，研究起步较晚。\cite{2019}发布了室外实拍浓雾数据集。\cite{2020NH-HAZE}发布了室外实拍非均匀数据集。\cite{NDIM}提出了NDIM算法，该算法在估计入射光的颜色特征后进行颜色校正步骤。 \cite{NHRG} 区分大气光、雾霾光、辉光和不同颜色的光源，提出了一种基于辉光特殊处理和夜间不同光源识别的算法。\cite{ancuti}等人提出了一种多尺度的人工光源补丁金字塔网络来适应夜雾环境。 \cite{MRP}认为夜间图像各颜色通道的局部最大强度主要由环境光照贡献，提出最大反射率先验的MRP算法。\cite{OSFD}基于场景几何，然后对光线和物体反射率进行二维模拟。提出了新的雾霾渲染图像提出了新方法和基准测试方法。
Compared to normal image dehazing, the thick haze, non-uniform haze, and night scene haze are more complex, and the research started later. \cite{2019Dense} released a real-world dataset of thick haze outdoors. \cite{2020NH-HAZE} released a real-world dataset of non-uniform haze outdoors. \cite{NDIM} proposed the NDIM algorithm, which performs colour correction after estimating the colour features of the incident light. \cite{NHRG} distinguished between atmospheric light, haze light, glow light, and different types of sources of light, and proposed an algorithm based on glow special processing and night-time different source recognition. \cite{ancuti} proposed a multi-scale artificial light patch pyramid network to adapt to night-time haze environments. \cite{MRP} believed that the local maximum intensity of each colour channel in night-time images was mainly contributed by environmental lighting, and proposed the MRP algorithm with maximum reflection first-order prior. \cite{OSFD} based on scene geometry, then simulated the light and object reflection rates in two dimensions. A new method and benchmark testing method for haze rendering images were proposed.

\subsection{Segmentation and large language model}
%不同于去雾是一项计算机视觉底层任务，语义分割是一项计算机视觉高层任务，其目标是将图像中的每个像素分类为一个类或对象。目标是生成图像的密集像素分割图，其中每个像素都分配给特定的类或对象。 此任务的一些示例基准是 Cityscapes、PASCAL VOC 和 ADE20K。 模型一般使用平均交并比（平均 IoU）和像素精度指标等进行评估。
Unlike image dehazing, which is a low-level task in computer vision, semantic segmentation is a high-level task in computer vision to classify each pixel in an image into a particular class or object. The goal is to generate a dense pixel segmentation map, where each pixel is assigned to a specific class or object. Some examples of benchmark datasets for this task include Cityscapes \cite{cityscapes}, PASCAL VOC \cite{voc}, and ADE20K \cite{ade20k}. Models are typically evaluated using metrics such as average intersection over union (average IoU) and pixel accuracy measures.

% 最近部分研究将图像编码的模型和大型语言模型关系。利用互联网上的大规模数据集预训练的大型语言模型正在通过强大的零样本和少样本彻底改变Natural Language Processing(NLP)泛化\cite{10languege}. 这些模型可以泛化到超出超过数据分布的特性。这些模型在数据集缺少的情况下甚至一定概率战胜端到端训练的微调模型\cite{21palm}。 经验趋势表明这种行为随着模型规模的增加而改善，同时数据集大小和训练模型尺度也会明显影响模型的性能。将成对的文本和来自网络的图像对齐是其中优秀的工作，例如CLIP\{82}和 ALIGN\cite{55}使用对比学习训练对齐两者的文本和图像编码器方式。 经过改进后可以对新图像场景和数据的零样本泛化。这样的编码方法也有效地与启用底层图像任务的其他模块，例如图像生成中的DALL·E.\cite{83delle}。
Recently, some studies have explored the relationship between image coding models and large language models. Pretrained large language models on massive datasets available online have revolutionized the generalization of natural language processing (NLP) \cite{10language}. These models can generalize to properties beyond the distribution of the data. These models can even defeat end-to-end trained fine-tuning models with a certain probability when there is a lack of dataset. Empirical trends indicate that this behavior improves as the size of the model increases, and the size of the dataset and the scale of the training model also significantly affect the performance of the model. Aligning pairs of text and images online is an excellent work, for example, CLIP \cite{82} and ALIGN \cite{55} use contrastive learning to train text and image encoders that align both pairs. After improving, such encoding methods can generalize to zero-sample scenarios for new image contexts and data. Such coding methods also effectively collaborate with other modules that enable fundamental image tasks, such as image generation by DALL·E \cite{83delle}.

\subsection{The haze effects on semantic segmentation}
%之前有一些研究关注了图形去雾和图像分割的关系。有雾的图像会造成现有模型检测分割能力的退化。
There have been previous studies that focus on the relationship between image dehazing and image segmentation. Haze can cause degradation in the segmentation capabilities of existing models.

%一部分研究关注图像去雾之后能提升多少网络的分割性能。图像通过去雾管道之后图像的分布更接近正常分割训练的数据，所以能提升分割和检测的能力。\cite{aodnet}就研究发现通过去雾的图像能获得更好的检测分割结果。
Some research focuses on the segmentation performance of the network can be improved after image dehazing. After passing the dehazing pipeline, the distribution of the image is closer to the distribution of the normal segmentation training dataset, thereby improving segmentation and detection capabilities. \cite{aodnet} found that dehaze images provided better detection and segmentation results. 

%另外一些研究者希望直接通过检测分割网络的改进来应对雾霾的图像退化。 FIFO\cite{fifo}提出一种区域映射的过滤器来提高雾霾场景下的图像分割能力。\cite{SFSU}和\cite{SUOF}分别提出了一种通过合成雾霾管道的方法来训练雾霾场景下的分割网络。
Other researchers hope to adapt the data environment of fog conditions by directly improving detection and segmentation networks. \cite{fifo} proposed a region mapping filter to improve the segmentation capabilities of images in haze-contaminated scenes. \cite{SFSU} and \cite{SUOF} respectively proposed a method of training segmentation networks in haze-contaminated scenes by synthesizing haze-contaminated pipelines.

\section{Methodology}
\subsection{Large model can overcome haze degradation}
% 我们认为大模型会涌现意想不到的能力。例如语言大模型经过大量训练可以进行对话等文本交互。大量数据和大尺寸的分割网络由于其参数足够多，并且参与的图像颜色和对比度分布比较广，我们认为其可以产生一定的雾霾域适应性，就如同The haze effects on Semantic Segmentation中讨论的一样。
%现有的分割大模型和去雾模型在模型参数和训练数据集上都存在几个数量级的差距。按照图像数量比较，分割大模型的数据量比去雾数据多四个数量级。同时分割大模型的参数量比去雾网络多两个数量级。具体对比如图\cref{FIG:1}
%大模型是的抗雾霾能力是自动获得的，不是人为特殊干预的。如果大模型能在一定的雾霾条件下进行精确分割，这将帮助小型去雾网络使用各种类型的雾霾。
%We use the Segment Anything model \ref{} as our large segmentation model. Its training data volume exceeds 1 billion tokens, and its network weight exceeds 2 GB.我们使用segment anything model \ref{}作为我们的分割大模型。其训练数据量超过1billion,网络权重超过2GB.
We believe that large-scale models can emerge with unexpected capabilities. For example, \cite{gpt3} generative pre trained transformer (GPT) trained extensively can perform text interaction such as dialogues. Due to their sufficient parameters and wide distribution of image color and contrast participation, large-scale segmentation networks can potentially achieve some degree of domain adaptability to haze. As the previous paragraph mentioned, other researchers have used methods such as image dehaze or domain adaptation to achieve the same results. However, the anti-haze capabilities of large models are automatically obtained, rather than manually optimized. The existing segmentation large models and dehaze models exhibit several orders of magnitude differences in their model parameters and training datasets. In terms of image numbers, the dataset size of segmentation large models is four orders of magnitude larger than that of dehaze models. Meanwhile, the parameter size of segmentation large models is two orders of magnitude larger than that of dehaze networks. The detailed comparison is shown in Fig.\ref{p1pipe}. We use the segment anything model (SAM) \cite{SAM} as our large segmentation model. Its training data volume exceeds 1 billion masks, and network parameters are close to 3GB, and the occupied video memory is close to 50GB when tested on the dehaze dataset. If large models can perform precise segmentation under haze conditions, this can assist small dehaze networks in various types of haze. 

%雾霾，图像，分割和纹理的关系。图中展示了不同类型雾霾场景下，对应的雾霾图，分割图，残差图和无雾图。分割结果可以在各种雾霾条件下将类似的雾霾退化区域分开，并保持基本的纹理区域边缘。图中的括号是图像在数据集当中的位置。
\begin{figure*}
        \centerline{\includegraphics[width=1\textwidth]{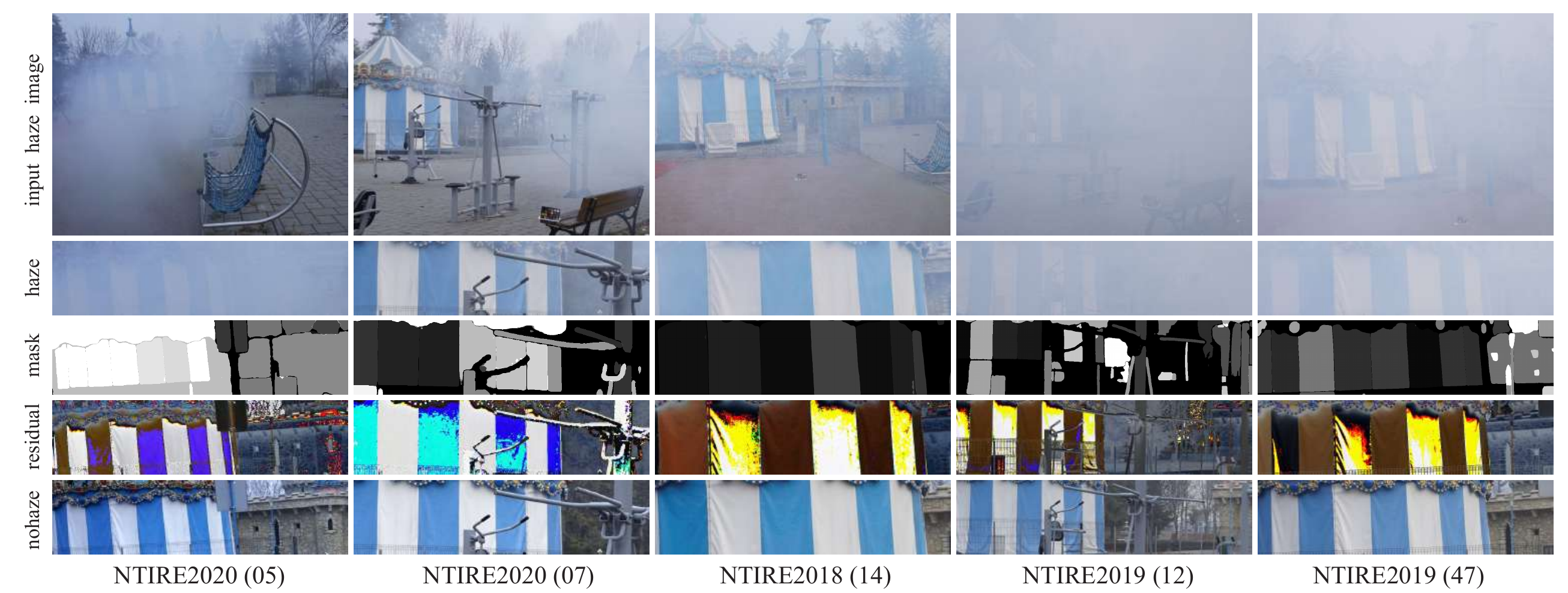}}
	\caption{\textbf{The relationship between segmentation and texture in haze images.} The figure displays different types of haze scenarios, input haze images, segmentation masks, residuals, and no haze images. The segmentation result can separate similar haze degradation areas under different haze conditions. The numbers in parentheses in the figure indicate the image position of the dataset.}
	\label{p6}
\end{figure*}

\subsection{Haze, image texture and segmentation}
% 雾霾 纹理和分割的关系
The traditional dehaze formula is as follows, 
\begin{equation}
I(x) = R(x)t(x)+L(x)(1-t(x)), 
\label{eq1}
\end{equation}
where $x$ is the position of the pixel, $I(x)$ is the signal received by the camera pixel, $R(x)$ is the signal emitted by the object itself, $L(x)$ is the atmospheric global illumination, and $t(x)$ is a transmission rate. The transmission rate formula is as follows,
\begin{equation}
t(x)=e^{-\beta \cdot d(x)},
\label{eq2}
\end{equation}
where $d(x)$ is the distance from the object to the camera, $\beta$ is the attenuation coefficient, and $e$ shows that the attenuation is in exponential.

%但是很多雾霾场景并不是均匀的,透过率也和深度不是线性关系。因此$t(x)$transmission rate的变化较为复杂。同时$L(x)$也受到雾霾本身性质和照射到雾霾的影响。雾霾的多变环境容易导致Eq.\cref{eq2}可以往往失效，传感器输入的$I(x)$往往很难直接线性计算出$R(x)$。

However, many haze images are not homogeneous and the transmittance is not linearly related to depth, the $t(x)$ is more complex. Additionally, $L(x)$ is influenced by the properties of the fog and the light. The variable environment of fog makes Eq.\ref{eq2} often fail, and it is difficult to directly calculate $R(x)$ from the input of the sensor $I(x)$ in a linear manner.

% 这里没有太看懂，R_{mask2texture}表示从分割的mask到无雾图像的转移矩阵？就是从一个0-1的mask到一个自然图像的变换矩阵？这样把segmentation和自然图像联系起来感觉有些强扭。

%但是大模型如果可以输出精确的分割结果，那么公式\cref{eq1}可以变成公式\cref{eq3}去. 由于我们最终需要的是$R(x)$.一旦有了$R_{segmask}(x)$的引导，从公式\cref{eq3}去拟合计算$R_{mask2haze}(x),再去通过公式Eq.\cref{eq4}去计算$R(x)$会比没有引导直接计算要简单。
If a large model can output precise segmentation results, Eq.\eqref{eq1} can be transformed into Eq.\eqref{eq3}.

\begin{equation}
I(x) = R_{mask}(x)R_{mask2texture}(x)t(x)+L(x)(1-t(x)), 
\label{eq3}
\end{equation}
% 其中 $R_{mask}$ 表示理想条件下大分割模型输出的分割mask, $R_{mask2texture}$表示从分割mask到无雾图像的转移矩阵。
where the $R_{mask}(x)$ represents the segmentation mask output by a large-scale segmentation model under ideal conditions, while $R_{mask2texture}(x)$ represents the transition matrix from the segmentation mask to the no haze image. Since we ultimately need $R(x)$, once we have the guidance of $R_{mask}(x)$, it will be easier to fit and calculate $R_{mask2texture}(x)$ using Eq.\eqref{eq3}. Then calculate $R(x)$ using Eq.\eqref{eq4},

\begin{equation}
R(X) = R_{mask}(x)R_{mask2texture}(x).
\label{eq4}
\end{equation}

% 上述是基于公式的理解，也可以从主观上理解分割信息的意义。不同物体往往具有不同的纹理，不同的纹理在雾霾的作用下会有不同的退化结果。分割掩膜提供了纹理边缘信息，并将类似退化的部分分割成一起，从而能更好的指导小网络学习类似退化过程。
In addition to relying on the dehaze formula, we can also subjectively understand the relationship between image segmentation and image texture degradation under fog conditions. As shown in Fig.\ref{p6}, different objects often have different textures, and different textures will suffer from different degeneration under the effect of fog. The segmentation mask provides texture edge information and segments similar degeneration parts together, thus better guiding small networks to learn similar degeneration processes.

% 灰度编码
\subsection{Grayscale coding of segmentation results}
%分割模型输出的是图像中像素对应分割的数字。其他分割项目中也将分割结果通过红绿蓝紫等颜色进行可视化，不同类型的图像用不同的颜色表示。分割成彩色的图像会占用三个通道，增加后续去雾小模型的计算开销，所以我们将不同分割类型的结果用放在一个灰度通道上。将不同类型的物体表示成0-255分布的灰度图像，如图\ref{p1pipe}(a)部分所示。由于大模型能够分割大量的物体，一张图像能被合理分割的图像数量很多。根据我们对于去雾数据集的实验，一张图能最多能分割132类型对象，同时大部分图像的分割数量往往在几十到255的一半之间。因此，我们提出了灰度编码方式，如算法\ref{Gray}所示，将分割结果变成灰度图像。有以下几个优势,第一个尽量占满1-255的灰度空间(0是没有分割的结果），从暗到亮可以明确每个分割结果的输出顺序。同一张图像分割结果越亮的区域越多，说明分割数量越多，网络性能更好，用灰度编码能更主观感受分割效果，如图\ref{}。

\begin{figure*}
 
        \centerline{\includegraphics[width=1\textwidth]{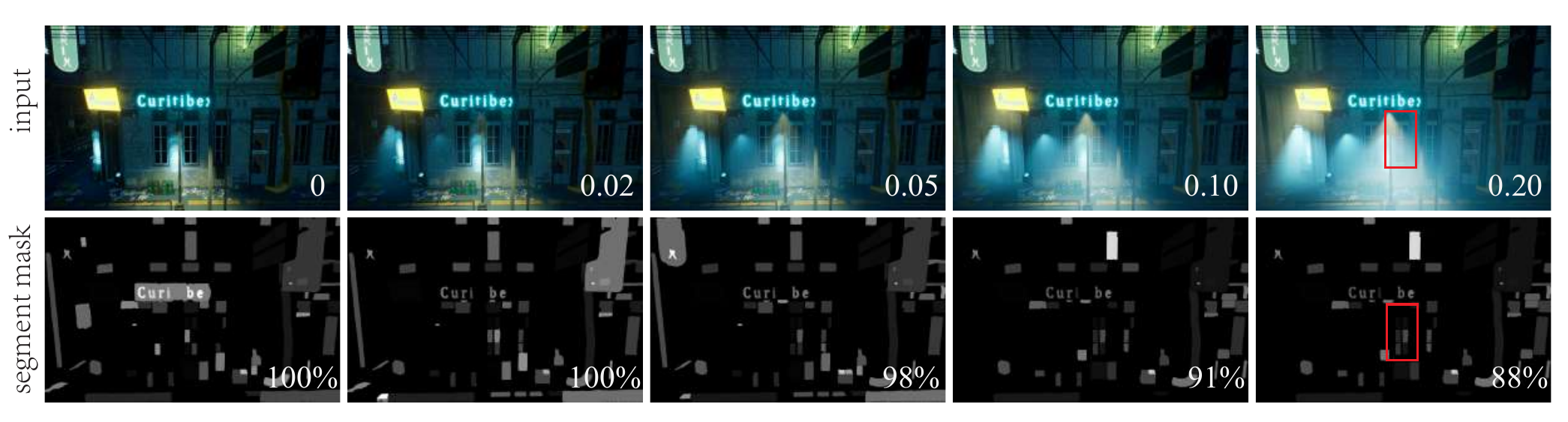}}
	\caption{\textbf{Different concentrations of haze and the performance of segmentation networks.} The images in the top row are input images with haze, and the numbers in the lower right corner are the fog concentration. The images in the bottom row are gray scale masks obtained after passing the segmentation network, and the numbers in the lower right corner are the percentage of segmentations compared to the no-haze case. As the fog concentration increases, the order of segmentation changes, and the mask that is segmented last is brighter. It is worth noting that in the red box, even if the fog is not evenly distributed, the model can still correctly segment each window position. Large-model segmentation networks can provide texture guidance for image dehazing. }
	\label{p7hazeadd}
\end{figure*}

The output of the segmentation model is a digital number corresponding to the segmentation of each pixel in the image. In other segmentation projects, the segmentation result is often visualized by converting the image to color, with different colors used to represent different types of segments in an image. Converting segments in an image to color requires three channels, which increases the computational complexity of subsequent small models for dehazing. Therefore, we propose to combine segmentation on a grayscale channel as shown in Fig.\ref{p1pipe}(a). Since large segmentation models can segment all objects in the image, there often exist many objects in an image. As a result, an image is often divided into many parts in segmentation result. According to our experiments on a dehazing dataset, an image can be segmented into more than 130 parts at most, while the majority of images have a segmentation number between 30 and 127 (half of 255). We propose a grayscale coding method, shown in Algorithm \ref{Grayscale coding of segmentation}, which converts the segmentation result into a grayscale image. There are several advantages to this method. First, try to fill the 1-255 grayscale space (0 is the result of no segmentation). From dark to bright, it can be clearly understood the output order of each segmentation result. There are brighter areas in the segmentation results of the same image, indicating more segmentation, and the network performance is better. Using grayscale coding helps to feel the segmentation effect more subjectively, as shown in Fig.\ref{p7hazeadd}.

%图中的表示了我们实验对比的不同方法的网络结构。网路结构的每一个色块都和Fig.\ref{}当中的一一对应的。
\begin{figure*}
        \centerline{\includegraphics[width=1\textwidth]{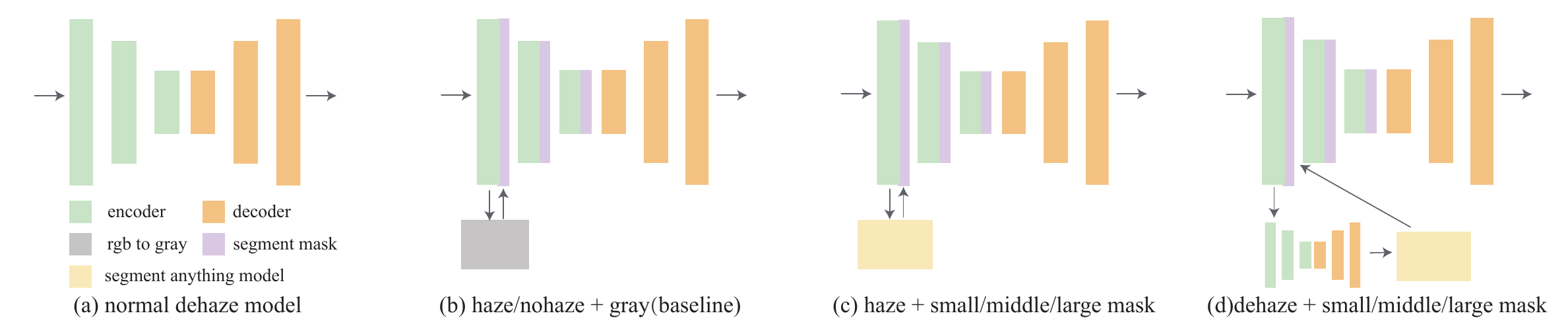}}
	\caption{\textbf{Network structure of image dehaze model with mask}. The figure shows the network structure of different methods compared in our experiments. Each color block of the network structure has a one-to-one correspondence with Fig.\ref{p1pipe}.}
	\label{p3netpipeline}
\end{figure*}

\begin{algorithm}[t]
    \caption{ Grayscale coding of segmentation}
    % 输入 ：分割网络的输出结果 其中包括分割类型，分割像素分布the 数据
    % 输出 ：灰度编码的图像 
    \label{Grayscale coding of segmentation}
    \KwIn{The output results of the segmentation network $masks$, including segmentation types $id$ and segmentation pixel distribution data $area$.} 
    \KwOut{Grayscale coding segment mask $segmask$} 
 
    \For{ $id$, $area$ in \textbf{enumerate}($masks$)}{
        % 分割结果矩阵图像化
        $area$ becomes a matrix with a segmentation result of 1\;
        \If {$id$ < 127}{
            $segmask = segmask + area\times2\times(id+1)$\;
            }
        \Else{
            $segmask = segmask + area\times(2\times(255-id)-1)$\;
            }
        }
        $segmask$ from matrix to single channel image\;
    \Return $segmask$\;
     
\end{algorithm}

\subsection{Image dehaze model with segmask}
% 常用的去雾模型往往输入和输出都是三通道的，输入的有雾图像RGB三通道，输出是去雾图像RGB三通道。为了能让模型能学会灰度编码的分割掩膜，我们需要将输入通道变成四通道的，增加一个灰度通道，放入分割掩膜。输出通道不发生变化，依然是三通道的。这样网络结构中的encoder将会变的多一个通道，整体膨胀，decoder部分不发生变化，大概网络架构如图\ref{}(c)所示。值得注意的是两者的encoder并不是分开的，只是通道不一样。
Common dehazing models typically have an input and output of three channels. The input fog image has RGB three channels, and the output is the dehazed image has RGB three channels. To enable the model to perceive grayscale encoded segmentation masks, we need to expand the input channels to four channels by adding a grayscale channel and putting the segmentation masks in the new grayscale channel. The output channels remain at three channels. This causes the encoder part of the network to have an additional channel and expand in size, while the decoder part remains unchanged, as shown in Fig.\ref{p1pipe}(c). It is important to note that the two encoders are not separate, but rather have different channels.

% 这种扩通道的方法绝大部分去雾网路都能使用，但是增加通道的之后参数会上升，原有的网络结构和改进后的结构参数并不相同，直接进行比较并不公平。为了去除参数变化的影响，单纯比较分割模型带来的优势，我们我们分别将RGB通道直接生成的灰度图像送入新的灰度通道作为基准。由于雾霾的浓度和分布不均匀，不同数据集难度不同，这个基准也作为评价数据集本身难易的基准。我们将网络的encoder的第四个通道放入以下几种类型的数据：
This expansion channel method can be used by most dehazing networks, but after increasing the number of channels, the parameters increase, and the original network structure and the improved structure have different parameters, making direct comparison unfair. To remove the impact of parameter changes, we generated grayscale images from the RGB channels into the new grayscale channel as the baseline. Since the concentration and distribution of haze are not uniform, the difficulty of different dehazing datasets is different, and this baseline also serves as a baseline for evaluating the difficulty of the dataset itself. We placed the fourth channel of the network's encoder into the following types, all corresponding network structures are shown in Fig.\ref{p3netpipeline}.

% 我们将进行下面几类不同maskputting网络结构，其中gray是指rgb图像对应的灰度图。其中the small/middle/large是指SAM模型当中不同的模型大小。large是指vit_h,是SAM model的最大模型。middle是指vit_l,中型尺度。small是指vit_b,是SAM当中最小的模型。
We will carry out the following types of dehazing network structures with different masks, where gray refers to the grayscale image from the RGB image. Among them, small/middle/large refers to different model sizes in the SAM model. Large refers to vit\_h, which is the largest model of the SAM model. Middle refers to vit\_l, medium scale. Small refers to vit\_b, which is the smallest model in SAM.

\begin{itemize} 
\item \textbf{haze gray}:
% 直接将有雾图像的灰度图作为segmask，送入网络。是实验的基准。
Baseline. The grayscale of the haze image is directly sent to the network as a segmask.

\item \textbf{nohaze gray}: Directly send the grayscale image of the fog-free image as segmask to the network. Is the absolute ideal ceiling for experiments. Used to evaluate data and network capabilities.
% 直接将无雾图像的灰度图作为segmask，送入网络。是实验的绝对理想的天花板。用于评价数据和网络能力。

\item \textbf{haze + small/middle/large}:
% 有雾图像经过不同大小模型分割后的结果作为segmanot uniform
% 这里需要说明一下small/middle/large代表的是模型的size（下同）
After the fog image is segmented by different size models(small/middle/large), it is sent to the network as a segmask.

\item \textbf{nohaze + small/middle/large}: 
After the no-haze image is segmented by different size models, it is sent to the network as a segmask. Used in evaluating segmentation network performance and dehazing training.
% 无雾图像经过不同大小模型分割后的结果作为segmask，送入网络。用在评价分割网络性能和去雾训练当中。

\item \textbf{dehaze + small/middle/large}: Dehazing first, and the result of dehazing image segmentation by different size models is sent to the network as a segmask. Used in the actual dehazing test.
% 先进行去雾，去雾图像经过不同大小模型分割后的结果作为segmask，送入网络。用在实际去雾测试当中。

\end{itemize} 

%不同浓度雾霾与分割网络的性能。上面一排图像是送入的有雾图像，右下角数字是雾霾浓度。下面一排图像是通过分割网络后的灰度掩膜，右下角数字是分割的数量相比无雾的百分比。随着雾霾浓度的升高，分割的顺序会发生变化，编码时候越亮是越往后分割出来的。注意到红色的框内，即便雾霾非均匀将窗框分开，检测网络依然可以正确切分每个窗户的位置。大模型分割网络能给图像带来去雾后的纹理指导。

\section{Data Preparation}

%现有的去雾数据集有很多，我们选择数据需要覆盖室内和室外，浓雾，薄雾和非均匀雾霾各种场景。这样才能检测我们的方法在不同雾霾下的效果。同时尽量追求实拍数据集，这样和真实场景更贴近。我们选择了RESIDE的STOS代表仿真室内薄雾，REVIDE代表室内浓雾。NTIRE2018I和NTIRE2018O代表了室内和室外实拍的雾霾。NTIRE2019代表了室外浓度，NTIRE2020代表了室外非均匀雾霾。

There are many existing dehaze datasets available, and we need to choose data that covers various scenarios such as indoor and outdoor, thick fog, thin fog, and non-uniform fog. This will allow us to test the effectiveness of our method under different fog conditions. At the same time, we should strive to use real-shot datasets as much as possible to achieve a closer match to real-world scenarios. We have chosen RESIDE to represent simulated thin fog outdoors, and REVIDE \cite{REVIDE} to represent thick fog indoors. NTIRE2018I \cite{2018I} and NTIRE2018O \cite{2018O} represent real-world fog captured indoors and outdoors, while NTIRE2019 \cite{2019Dense} represents outdoor thick fog and NTIRE2020 \cite{2020NH-HAZE} represents outdoor non-uniform haze.

% 由于不同类型的雾霾，会导致不用雾霾退化的关系。我们通过不同的类型雾霾数据代表不同雾霾场景来分析大模型对于小模型的指导能力。从而了解分割mask对于不同类型雾霾退化的关系。
 
Due to the existence of different types of haze, there is a relationship between the deterioration of different haze types. We analyze the guidance ability of large segment models for small dehaze models based on different types of dehaze data. The dehaze result in different fog scenarios can show the relationship between the degradation of different fog types and the segmentation mask.

% 同时为了检测非均匀雾霾和不同雾霾浓度对于分割网络的影响，我们使用了虚幻5引擎来渲染生成不同浓度的雾霾来评价大模型雾霾下的分割能力。仿真图像中雾霾浓度是基于体积雾来渲染的，随着体积雾雾霾浓度的强度变化，图像当中的雾霾逐渐增加。
To compare the impact of non-uniform haze and different haze concentrations on the segmentation network, we used Unreal Engine 5 \cite{UE5} to render and generate different concentrations of fog to evaluate the segmentation ability of the network under haze conditions. The simulation images were rendered based on the volume fog \cite{UE5fog}, and the strength of the fog concentration varied with the intensity of the volume fog. As the fog concentration increased, the image became increasingly polluted.

%不同去雾数据集不同类型分割掩膜结果。

%从上倒下分别展示了不同去雾数据集当中有雾图像的分割结果。挑选的图片和下图fig5的去雾结果是一一对应的。从左到右分别是有雾输入图像，不同尺寸模型的分割结果和先去雾之后不同尺寸模型的分割结果。

\begin{figure*}
 
        \centerline{\includegraphics[width=1\textwidth]{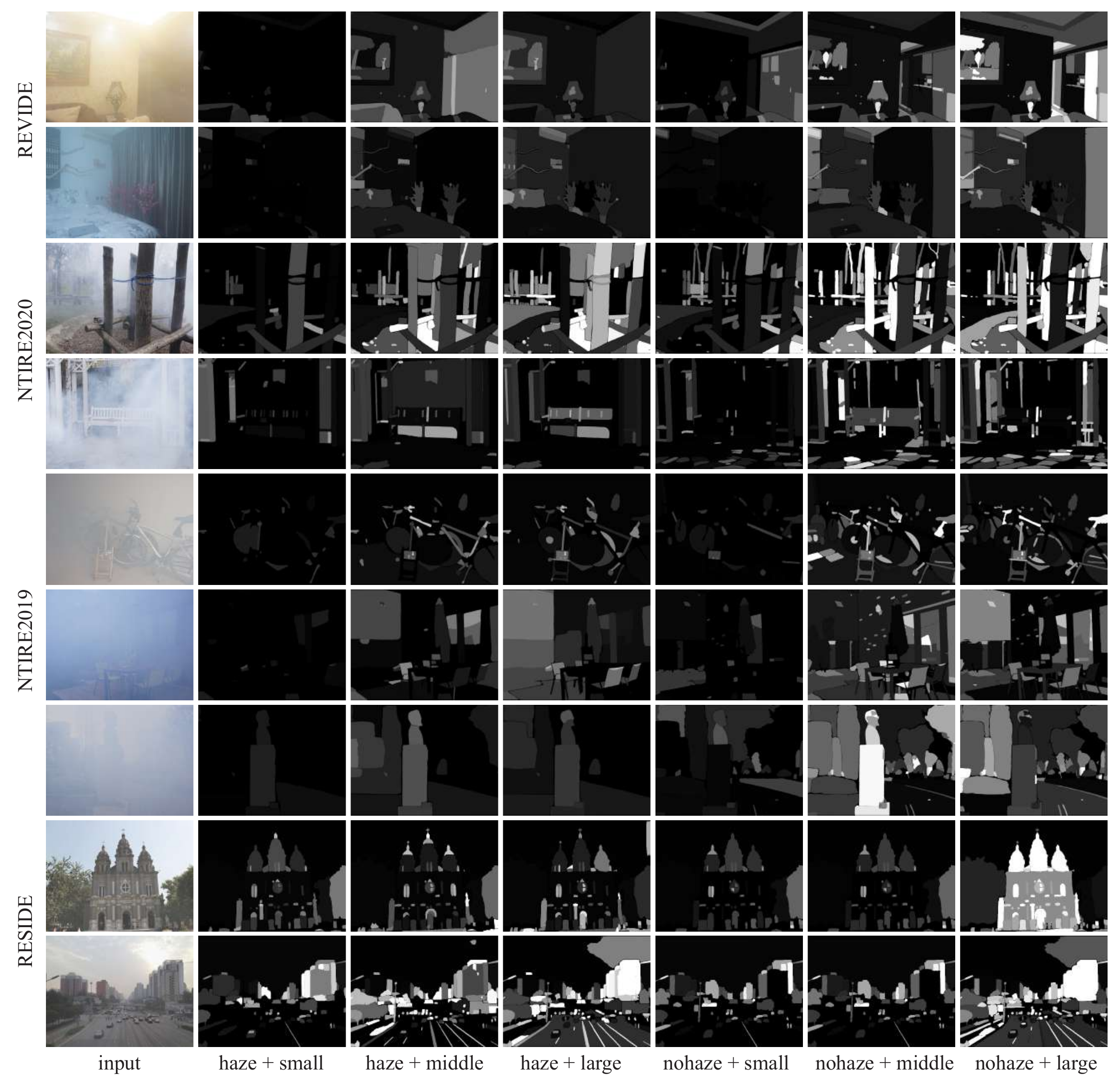}}
 
	\caption{\textbf{Results of different types of segmentation masks for different dehazing datasets.} The figure presents the segmentation results of hazy images in different dehazing datasets from top to bottom. The selected images correspond to the dehazing results in Fig.\ref{P5result abla}. The results from left to right are the hazy input images, different size models segmentation results with haze and dehazing.}
	\label{p4segresult}
\end{figure*}

\section{Experimental Assessment}
% 雾霾对检测的影响
\subsection{The haze effects on segmentation}
% 一般的检测分割数据集并不会使用有雾霾的数据集来进行训练。SAM大模型训练的数据集中也没有关于雾霾的数据。所以雾霾的退化会对分割网络的性能造成负面的影响。
The general detection and segmentation datasets do not use datasets with haze for training. The dataset used for the SAM large model training does not have any data related to haze. Therefore, the degradation caused by haze will have a negative impact on the performance of the segmentation network.

% 我们定性和定量的衡量了雾霾浓度浓度对于分割网络性能的负面影响。由于现有的去雾数据集没有不同浓度和非均匀数据，我们使用虚幻引擎来渲染不同浓度的雾霾数据。我们将这些不同浓度的数据分别通过分割大模型。将没有雾霾的图像作为基准进行测试。雾霾浓度从0增加到0.2,相当于从无雾霾变为很浓的雾霾，检测出来数量逐渐从基准数量的100%逐渐退化都88%。如图\ref{p7hazeadd}红框区域所示，由于大模型的能力，在非均匀浓雾的恶劣情况下，大体量的分割模型依然可以正确的分割窗户的每一个玻璃。
We quantitatively and qualitatively measured the negative impact of haze concentration on the performance of the segmentation network. Since the existing dehaze datasets do not contain different concentrations and non-uniform haze images, we used Unreal Engine 5 \cite{UE5} to render different concentrations of smog data. We passed these different concentrations of data through the segmentation large model separately. The baseline was an image without fog. As the fog concentration increased from 0 to 0.2, which corresponds to going from no haze to very thick haze, the detection rate gradually decreased from 100\% to 88\%. As shown in the red box area in Fig.\ref{p7hazeadd}, due to the ability of large models, in severe conditions of uneven thick fog, a large-scale segmentation model can still correctly segment each glass of the window.

\begin{table*} 
 \caption{Quantitatively compare the segmentation performance of models of different sizes under different datasets}
 \label{tab_seg_modelsize}
\centering	
\begin{tabular}{lcccccc}  
\toprule   
 Method  &  nohaze small &  hurtle & haze middle & nohaze large & haze large   \\  
\midrule   
RESIDE  & 80 (61\%) & 77 (58\%)& 131 (99\%) & 127 (96\%)  & 132 (100\%) & 130 (98\%) \\
REVIDE  & 55 (65\%)& 50 (59\%)& 82 (96\%)  & 74 (87\%) & 85 (100\%) & 75 (88\%)  \\
NTIRE2018O  & 53 (76\%)& 51 (73\%) & 69 (98\%)  & 45 (64\%)  & 70 (100\%) & 47 (67\%) \\
NTIRE2018I  & 52 (54\%) & 48 (50\%)  & 88 (92\%) & 87 (91\%) &  96 (100\%) & 90 (94\%)\\
NTIRE2019  & 32 (49\%) & 8 (12\%)  & 61 (94\%) & 15 (23\%) & 65 (100\%) & 16  (25\%) \\
NTIRE2020  & 47 (52\%) & 33 (36\%)  & 94 (103\%) & 59 (65\%)  & 91 (100\%) & 57 (63\%)\\
  \bottomrule  
\end{tabular}
\end{table*}

% 模型大小的影响
\subsection{Model size on haze image segmentation}

% 我们分别对比了不同尺度的模型对于雾霾的分割能力\ref{}。左侧从上到下代表了不同类型的雾霾，包括室内浓雾，室外非均匀雾霾，室外浓雾，室外薄雾。随着分割模型尺度的增大，模型风格数量和准确度越来越高。图像当中出现了越白的mask代表了分割出来数量越多。右边代表了无雾霾时候不同尺寸雾霾的分割能力。对比主观分割结果我们可以发现，对于RESIDE这种较为简单的雾霾数据有雾的分割结果往往比无雾还要好。其他较为困难的雾霾场景，往往是模型尺寸越大，能分割的越好，值得注意的是，当分割模型参数量变大一级的时候，往往能直接抵消雾霾的副作用。例如haze+middle的效果往往比nohaze+small效果好。这从主观上表明了分images的抗雾霾能力在各种复杂雾霾场景中的作用。至于在RESIDE中分割大模型效果较差，主要是因为small模型已经能在较轻的雾霾场景下进行有效的分割了，这种情况在后面对于去雾结果对比中也再次出现。
We compared the segmentation ability of different size models for different haze datasets in Fig.\ref{p4segresult}. The left side represents different types of haze, including indoor thick fog REVIDE, outdoor non-uniform fog NTIRE2020, outdoor thick fog NTIRE2019, and thin fog RESIDE. As the size of the segmentation model increases, the number of segmentation categories and accuracy will also increase. A mask that appears more white in the image represents more segments that can be extracted. The right side represents the segmentation ability of different size segmentation without haze. We found that for RESIDE, a relatively simple haze data set, the segmentation results with fog are often better than those without fog. In more difficult haze scenarios, the larger the model size, the better the segmentation accuracy. It is worth noting that when the parameter number of the segmentation model increases by one level, it often directly offsets the effects of haze. For example, the effect of haze + middle is often better than nohaze + small. This subjectively demonstrates the role of the emerging anti-haze ability of the segmentation network in various complex haze scenarios. For the large model performance on RESIDE is poor, mainly because the small model can already perform effectively in less severe haze scenarios, which is also repeated in the comparison of dehaze results later.

%使用大模型来加速去雾网络datasets图表示了不同类型数据集在使用和不使用大模型掩膜训练的收敛结果。下图表示了相同数据集，在使用不同尺寸大小分割大模型掩膜训练的收敛结果。
\begin{figure}
        \centerline{\includegraphics[]{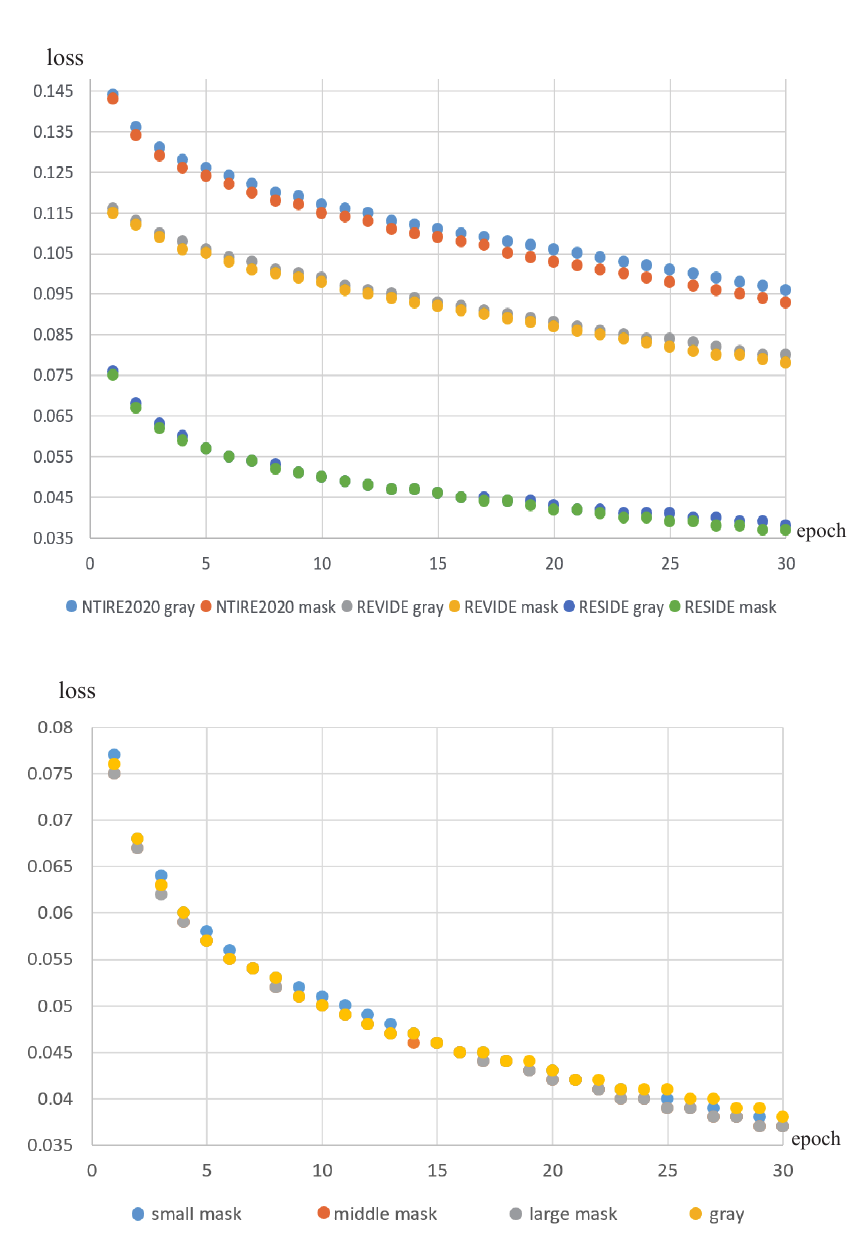}}
	\caption{\textbf{ The use of large models can accelerate the dehaze model training.} The above figure shows the convergence results of different data sets when trained with and without large segment model masks. The below figure shows the convergence results of the same data set when trained with different sizes of large segment model. }
	\label{p8loss}
\end{figure}

% 我们也统计了所有去雾数据集在不同大小分割模型下的定量表现，如表\ref{}. 所有的数据是分割网络分割整个去雾数据集之后每张图能分割数量的平均数。总体来说，简单的雾霾场景，小分割模型足够升任。中等难度的场景，模型尺度变大可以客服雾霾的副作用。及其困难的雾霾场景，模型尺度变大可以缓解雾霾的副作用。简单的RESIDE数据集，由于雾霾变化比较简单，有雾和无雾对分割能力影响非常的小。同时large尺度模型相比于middle尺度模型也没有什么明显提升。REVIDE, NTIRE2018O和NTIRE2018I都保持了类似的分割结果，就是雾霾会对分割造成一定的影响，但是分割网路的模型变大能克服雾霾的负面效果。NTIRE2019和NTIRE2020却展示了这两个数据的特殊性。NTIRE2019是浓雾的数据集，有雾和无雾的区别非常夸张，无论尺度如何增加，分割模型也无法使用极其微弱甚至本来就不存在的信号。但随着网络尺度的增加，其对于分割的能力也能提升超过100%。NRIRE2020是非均匀的去雾数据集。非均匀的雾霾对于去雾小模型是一个挑战，然而对于分割大模型来说，非均匀的雾霾干扰几乎没有效果，如图\\ref{p4segresult}中的NTIRE2020. 同时对于非均匀雾霾，大模型直接分割的效果也超过了先去雾之后小模型分割的效果。
We also conducted quantitative performance statistics on all dehaze datasets under different size segmentation models, as shown in Table \ref{tab_seg_modelsize}. All data is the average number of segments per image after the segmentation network segments the entire dehaze dataset. In general, small segmentation models are sufficient for simple fog scenes. For moderate-difficulty scenes, increasing the model scale can overcome the side effects of haze. For extremely difficult fog scenes, increasing the model scale can alleviate the side effects of haze. The fog changes on the RESIDE dataset are simple. Simple haze cannot cause the degradation of large-model segmentation results. The large-model segmentation results of haze images and clear images are not different. At the same time, neither size of the segmentation large model shows a significant improvement in subsequent image dehazing. REVIDE, NTIRE2018O, and NTIRE2018I maintain similar segmentation results, indicating that fog can cause some impact on segmentation, but increasing the scale of the segmentation network can overcome the negative effects of fog. NTIRE2019 and NTIRE2020 show the uniqueness of these two datasets. NTIRE2019 is a dataset of dense fog, and the difference between the presence and absence of fog is very dramatic. No matter how the scale increases, the segmentation model cannot use extremely weak or even nonexistent signals. However, as the size of the network increases, some weaker but still detectable signals are still correctly segmented by the large model. Segmentation results can improve by 100\% or more. NTIRE2020 is an unevenly distributed dehaze dataset. Uneven fog is a challenge for small dehaze models, but for large segmentation models, uneven fog interference has almost no effect, as shown in Fig.\ref{p4segresult}. The gradient haze border does not affect the segmentation effect. At the same time, the performance of large models directly segmenting uneven fog is also superior to the performance of small models after dehaze processing.

% 不同类型分割掩膜指导下的小网络去雾结果。 
% 所有的图像都和前文的fig3一一对应，是前文图像的放大展示。从上到下是代表不同雾霾类型的不同数据集。从左到右是不同去雾结果，红色框是值得关注的地方。对于现有去雾网络很难处理好的场景，可以看到随着分割模型尺度的增大，雾霾去除的结果是越来越好的。
\begin{figure*}
        \centerline{\includegraphics[width=1\textwidth]{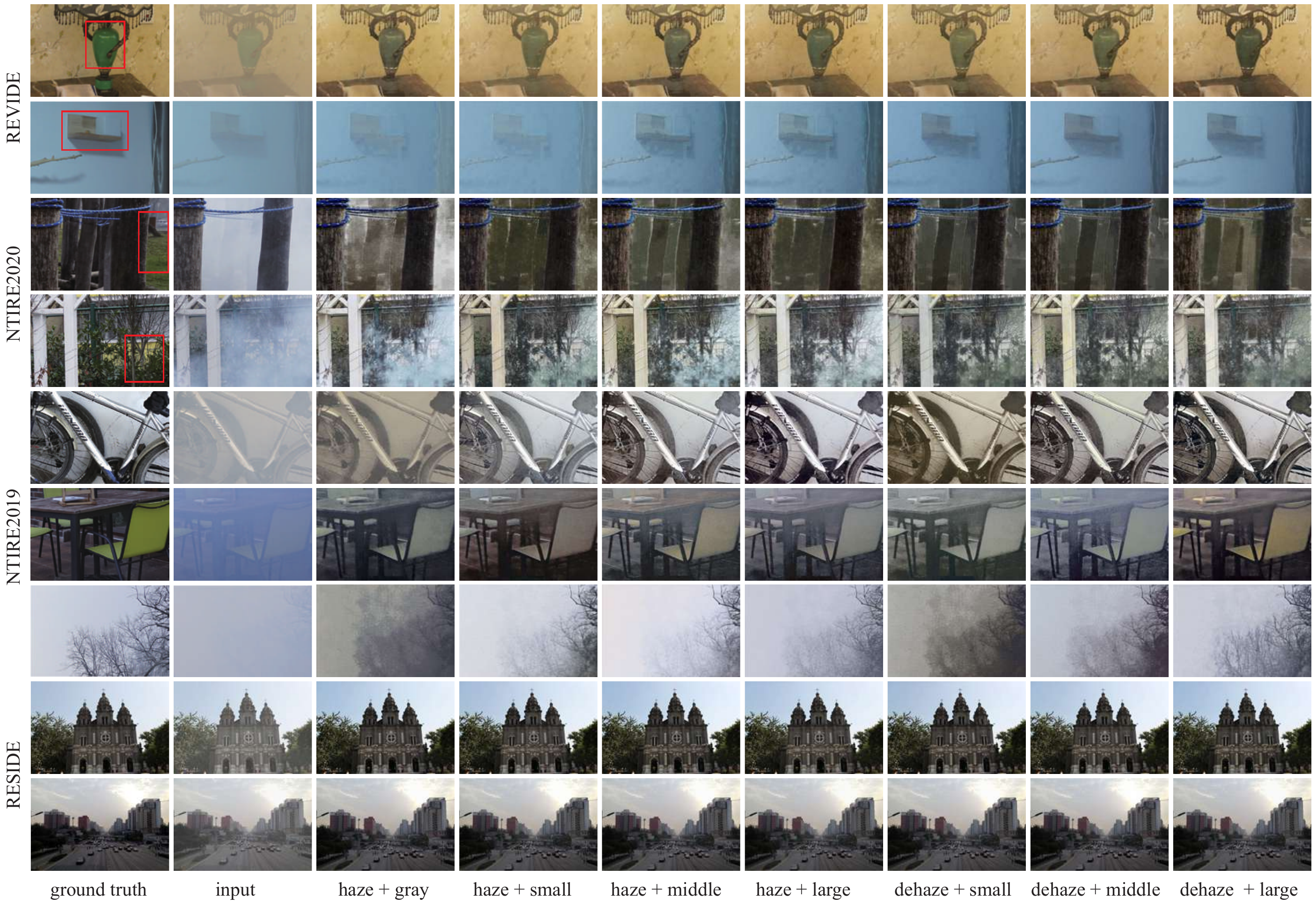}}
	\caption{\textbf{ Small network dehazing results guided by different types of segmentation masks.} All of the images correspond to Fig.\ref{p4segresult} in the previous paragraph, which is a magnified version of the previous image. From top to bottom, they represent different datasets of different types of fog. From left to right, they show different dehaze results. The red boxes highlight relevant areas that are worth attention. In scenes that are difficult for existing dehaze networks to handle, it can be seen that as the scale of the segmentation model increases, the quality of the dehaze results improves significantly.}
	\label{P5result abla}
\end{figure*}

\begin{table*} 
 \caption{Quantitative results comparing the universality of different datasets (\textcolor{olive}{best} and \textcolor{teal}{second best})}
 \label{tab_data}
\centering	
\begin{tabular}{lccccccc}  
\toprule   
 Dataset    & gray  &  small  &  middle  & large  \\  
\midrule   
RESIDE (haze)           &\textcolor{olive}{33.81}(0\%) / \textcolor{olive}{0.988}  & \textcolor{teal}{33.72}(1\%) / \textcolor{teal}{0.988}  & 32.46(17\%) / 0.988  & 32.65(14\%) / 0.982 \\
RESIDE (nohaze/dehaze)                   & \textbf{57.12 / 0.999}  & 33.67(16\%) / 0.988  & 33.70(13\%) / 0.988 &  32.84(12\%) / 0.988\\

NTIRE2018I (haze)     & 28.42(8\%) / 0.938    & \textcolor{olive}{29.05}(0\%) / 0.946  & 27.97(13\%) / \textcolor{olive}{0.961} & 27.73(16\%) / \textcolor{teal}{0.956}\\
NTIRE2018I (nohaze/dehaze)           & \textbf{53.38 / 0.998} & \textcolor{teal}{28.47}(7\%) / 0.914  & 27.35(22\%) / 0.951 & 27.31(22\%) / 0.929 \\

NTIRE2018O (haze)     & 26.03(1\%) / 0.795 &  26.06(1\%) / \textcolor{teal}{0.796}  & 26.08(1\%) / 0.794   & 26.01(2\%) / 0.794 \\
NTIRE2018O (nohaze/dehaze)             & \textbf{49.90 / 0.997} &  26.09(1\%) / 0.793 & \textcolor{teal}{26.14}(0\%) / 0.793 & \textcolor{olive}{26.15}(0\%) / \textcolor{olive}{0.796}\\

REVIDE (haze)         & 22.03(17\%) / 0.878  & 22.42(12\%) / 0.880 & 22.66(9\%) / 0.880  & 22.61(10\%) / \textcolor{teal}{0.883} \\
REVIDE (nohaze/dehaze)                 &\textbf{42.66 / 0.998} & 22.58(10\%) / 0.873 & \textcolor{teal}{22.86}(6\%) / 0.882 & \textcolor{olive}{23.41}(0\%) / \textcolor{olive}{0.888} \\

NTIRE2020 (haze)      & 19.90(13\%) / 0.658   & 20.52(5\%) / 0.652  & 20.54(5\%) / 0.652  &  \textcolor{teal}{20.69}(3\%) / \textcolor{teal}{0.659}\\
NTIRE2020 (nohaze/dehaze)              &  \textbf{50.11 / 0.999}   &  20.33(7\%) / 0.655  & 20.68(3\%) / 0.655  &   \textcolor{olive}{20.94}(0\%) / \textcolor{olive}{0.658}  \\

NTIRE2019 (haze)      & 16.46(17\%) / 0.415   & 17.15(8\%) / 0.578  & 17.32(6\%) / 0.575 & \textcolor{teal}{17.62}(2\%) / 0.575\\
NTIRE2019 (nohaze/dehaze)              & \textbf{49.51 / 0.999}   & 17.26(7\%) / 0.545 & 17.38(5\%) /  \textcolor{teal}{0.587} & \textcolor{olive}{17.84}(0\%) / \textcolor{olive}{0.587}\\

  \bottomrule  
\end{tabular}
\end{table*}

% 加速收敛
\subsection{Segmentation mask accelerates training}
% 由于分割网络能够涌现出抗雾性能，可以增强去雾的效果。同时我们还发现其可以提高雾霾网络训练的收敛速度，如图\ref{}所示。我们认为大模型产生的分割结果给去雾小网络提供了一个指导。大模型将相同的纹理变化区域进行了一个分割，帮助了去雾模型加速训练。我们的所有实验结构都显示出了这一特性。如图\ref{}上图中蓝红点所示,相比于其他两条曲线，红蓝点的loss收敛较慢，代表雾霾条件更复杂，更难处理的去雾数据集, 但segmentation mask 能加速的更明显。同时，这个加速效应即便是在较为简单的数据集上也微弱的存在。将Fig.\ref{}上图中的蓝绿线放大就得到了Fig.\ref{}中下图的结果。不同尺度的模型能产轻微的扰动，总的来说是大模型分割mask加速快于小模型快于没有分割的baseline.
Due to the ability of large model segmentation networks to emerge with anti-fog performance, they can enhance the effectiveness of dehaze. We also found the segmentation mask can improve the convergence speed of dehaze network training, as shown in Fig.\ref{p8loss}. We believe that the segmentation results produced by large models can guide small dehaze networks. Large models segment the same texture variations into different regions, helping dehaze networks accelerate training. All of our experimental structures demonstrate this property. As shown in the blue and red points in the upper figure of Fig.\ref{p8loss}, compared to the other curves, the loss of the blue and red points converges slowly, representing more complex and difficult to process dehaze datasets under fog conditions. However, red dots fall faster than blue dots, the segmentation mask can significantly accelerate training. Moreover, this acceleration effect is also present even on simpler dehaze datasets. Enlargement of the blue and green lines in the upper figure of Fig.\ref{p8loss} gives the result in the lower part. Different-scale segmentation models can generate slight perturbations, but in general, the segmentation mask of large models accelerates faster than small models and the baseline.

% 雾霾检测 循环检测的消融实验
\subsection{Comparison of qualitative dehazing results}
% 如图\ref{}展示了了部分去雾结果示意图。其中的图像都是fig.\ref{}当中一一对应的局部放大。对应区域的分割掩膜也可以在fig.\ref{}中对比查看。对于RESIDE这一类简单的数据集，现有的去雾网络可以轻松胜任，无论使用任何体量的大模型主观上去雾结果提升不大，如如图\ref{}最下面两行所示。对于浓雾和非均匀场景，我们的方法就能看到很明显的主观提升。随着分割掩膜的质量提升，去雾的质量也快速提升，无论是去雾的程度，还是边缘的细节都恢复的更好。REVIDE中，红框中的墨绿色花瓶和空调插座都因为有了纹理指引，纹理细节和颜色恢复的都更好。NTIRE2020当中由于非均匀雾霾容易造成去雾恢复的错误，例如fig.\ref{}第三行haze+gray树干无法保持相同的颜色。但是有了分割模型的指导，树干都能被正确的去雾。有了预先去雾步骤和分割之后，第三行NTIRE2020红框后面的纹理细节去雾网络也可以进行一定的恢复。在NTIRE2019三行展示图像当中，由于分割的指导，自行车车架，黄色的椅子，还有处理较为困难的树干枝杈。值得注意的是，一旦前置步骤去雾错误也会带来副作用，例如NTIRE2019第三行，dehaze+middle,一旦去雾错误成了haze+gray的样子，分割网络也会错误最终导致去雾结果的错误。但从视觉主观上来说，图像去雾结果的收益于大模型的抗雾分割能力。
Fig.\ref{P5result abla} shows an example of the results of dehazing by different masks. The images in the Fig.\ref{P5result abla} are the local enlargements of the corresponding regions in Fig.\ref{p4segresult}. The segmentation masks in Fig.\ref{p4segresult} can also be compared with the corresponding regions in Fig.\ref{P5result abla}. For simple datasets such as RESIDE, existing dehaze networks can easily perform well. However, when faced with thick fog and non-uniform haze scenes, our method shows significant subjective improvements. As the quality of the segmentation mask improves, the quality of the dehaze processing also rapidly increases. Both the color and the details of the edge are restored better. In REVIDE, the green vase in the red box and the air conditioning socket are better dehaze due to the segment mask guidance, resulting in better texture details and colors. In NTIRE2020, non-uniform haze is prone to causing errors in dehaze restoration, such as the trunk in Fig.\ref{P5result abla} third row being unable to maintain the same color by haze+gray baseline. However, with the guidance of the segmentation model, the trunk can be correctly dehaze. After the pre-dehaze step and segmentation, the texture detail de-fog network can also perform some restoration of the texture in the third row of NTIRE2020 red box. In the NTIRE2019 three rows of displayed images, due to the guidance of the segmentation model, the bicycle frame, the yellow chair, and the difficult to process branch nodes are also better resolved. It is worth noting that once the pre-de-fog step is incorrect, it can also have side effects. For example, in the third row of NTIRE2019 dehaze+small, if the pre-dehaze step is incorrect, like haze+gray, the segmentation network will also be incorrect and ultimately lead to incorrect dehaze results. However, from a subjective visual perspective, the dehaze results benefit from the anti-fog segmentation ability of large models.

\subsection{Comparison of quantitative dehazing results}
% 我们还定量比较了不同雾霾条件下，本文方法的提升效果。总的来说是和主观感受完全一致的,如表\ref{}所示。数据是按照baseline(haze+gray)从高到低排序的，代表了不同数据的难度。其中橄榄色是第一好。墨绿色是第二好。可以明显看有颜色的结果是对角线分布，对越难的数据我们的方法效果更好，提升更多，对于简答的数据，我们的方法就相对提升不大。对于简单的RESIDE来说完全没有提升，因为去雾小网络本身已经可以很好的完成去雾工作了。NTIRE2018I和NTIRE2018O相对中等，我们的方法能提升1%到8%但是后期的分割信息很容易饱和甚至产生副作用。而对NTIRE2020，NTIRE2019,REVIDE这类较难的场景，我们的方法就能提升13%-17%，并且分割模型尺度的增大对去雾的提升非常明确。
We also quantitatively compared the improvement effects of our method under different levels of fog. Overall, the results were consistent with subjective perceptions, as shown in Table\ref{tab_data}. The data was sorted from high to low based on the haze gray mask baseline's PSNR, representing the difficulty of different datasets. The olive color represents the best result, while the teal color represents the second-best result. It can be clearly seen that the color results are diagonally distributed, indicating that our method performs better and improves more for more difficult dehaze datasets and relatively little for easier dehaze datasets. For the simple RESIDE, there was no improvement because the dehazing small network itself can perform well. The NTIRE2018I and NTIRE2018O were relatively moderate, and our method could improve by 1\% to 8\%. For medium-level difficulty dehaze datasets, larger-sized segmentation networks quickly produce saturated effects and cannot continue to provide effective texture segmentation. For NTIRE2020, NTIRE2019, and REVIDE, which were more difficult scenarios, our method could improve by 13\% to 17\%. The increase in segmentation model scale can improve dehazing performance.

%同时我们还研究了大模型指导去雾的最理想效果，表\ref{}粗体部分所示。我们将没有雾的灰度图像作为mask送入网路进行去雾，网络能快速学会并接受。可见现有的大模型很大潜力去挖掘。这项数据也在下部分用来评价小网络的学习能力。
Additionally, we also studied the optimal effect of using large models to guide dehazing, as shown in the bold part of Table \ref{tab_data}. We applied a grayscale image without haze as a mask to the network for dehazing, and the network could quickly learn and accept it. It appears that there is much potential to explore with existing large models. This method is also used in the following section to evaluate the learning ability of different small dehaze networks.

% 小网络的学习能力
\subsection{Mask awareness of dehazing network}

% 这里想表达的可能不是mask learning而是小模型利用这个mask的能力，看看这个learning的词有没有什么更贴切的表述？

%模型尺寸对于分割大模型能力很关键，同时对于去雾小模型接受分割大模型的能力也很重要。虽然分割大模型能用将抗雾能力传递给去雾网络。但是不同尺度的去雾网络在相同的训练资源下，表现出了不同的学习能力. 如表格\ref{},10M,60M,100M代表了不同模型的尺度大小。

% 为了充分体体现网络的学习能力和各去雾先验无关，我们没有选用各种专门为去雾定制的模型例如dehazeformer和C2PNet,我们认为这些网路容易更适合某一些类雾霾场景，我们选择了图形恢复算法中通用不同尺寸模型作为代表，Unet\ref{}参数量10M，Uformer参数量60M,Restormer参数量100M.
To fully demonstrate the network's learning ability and independence from prior knowledge of dehazing, we did not choose various custom-designed models specifically for dehazing, such as Dehazeformer \cite{dehazeformer} and C2PNet \cite{c2pnet}. We believe these networks may be more suitable for certain types of hazy scenes. Instead, we selected general models for image restoration algorithms with different sizes, represented by Unet \cite{Unet}, Uformer \cite{Uformer}, and Restormer \cite{Restormer}.

%对于简单的去雾数据集，无论去雾网络的尺度大小，都无法展现分割mask的提升作用。对于之前讨论的分割掩膜能有明显提升的数据集，往往网络权重更大，更复杂，拟合能力越好的去雾小模型，越能学习大模型分割网络的透雾性能。
%同时更大尺度的去雾小模型能更好的感知到分割大模型的尺度提升。
%为了充分体现大模型和小模型之间的学习关系，前面的展示均基于最大尺度的去雾小模型来测试。

The size of the model is crucial for the ability of a large-scale segmentation model, and it is also important for a small-scale dehaze model to accept the ability of a large-scale segmentation model. Although large-scale segmentation models can transmit anti-fog ability to defog networks, different scales of defog networks exhibit different learning abilities under the same training resources. As shown in Table\ref{tab_network}, 10M, 60M, and 100M represent the scale of different models. For simple dehaze datasets, the improvement caused by the segmentation mask is not apparent regardless of the scale of the dehaze network. For more complex scenarios of haze, bigger size and more complex small-scale dehazing models that have better fitting abilities can often learn the anti-haze performance of large-scale segmentation networks. Furthermore, larger-scale dehaze models can better perceive the improvement of large-scale segmentation models. To fully reflect the learning relationship between large and small models, the previous demonstrations have all been tested using the largest-scale dehaze model.

\begin{table*} 
 \caption{Quantitative results comparing the universality of different networks (\textcolor{olive}{best} and \textcolor{teal}{second best})}
 \label{tab_network}
\centering	
\begin{tabular}{lcccccccc}  
\toprule   
 RESIDE  & nohaze & haze gray & haze small & de. small &  haze mid. & de. midd. & haze large & de. large \\  
\midrule   
Unet10M        & \textbf{41.79}&\textcolor{teal}{28.05}(2\%) &27.78(5\%)&27.65(7\%)&27.91(4\%)&\textcolor{olive}{28.22}(0\%)  & 27.43(10\%) &  27.32(11\%)  \\
Uformr60M       & \textbf{45.64}  & \textcolor{teal}{30.24}(4\%) & 29.55(13\%) & \textcolor{olive}{30.59}(0\%)  & 29.55(13\%) & 29.73(10\%) & 29.88(8\%)  & 30.11(6\%) \\
Restor.100M  & \textbf{57.12}  & \textcolor{olive}{33.81}(0\%) &  \textcolor{teal}{33.72}(1\%) & 33.67(2\%) & 32.46(17\%) & 33.70(1\%) & 32.65(14\%) &  32.84(12\%) \\
\midrule   
 NTIRE2019 & nohaze & haze gray & haze small & de. small &  haze mid. & de. midd. & haze large & de. large \\  
  \midrule  
Unet10M        & \textbf{25.49} & 14.15(1\%) & 14.11(1\%) & 14.11(1\%)  & 14.10(1\%) & 14.13(1\%) & \textcolor{olive}{14.21}(0\%)  &  \textcolor{teal}{14.17}(0\%) \\
Uformr60M   & \textbf{26.28}  & 15.04(5\%) & 15.19(3\%) & 15.27(2\%) & 15.32(2\%) & \textcolor{teal}{15.45}(0\%) & 15.42(0\%) & \textcolor{olive}{15.45}(0\%) \\
Restor.100M  & \textbf{49.51} & 16.46(17\%) & 17.15(8\%) & 17.26(7\%)  & 17.32(6\%) & 17.38(5\%) & \textcolor{teal}{17.62}(3\%) & \textcolor{olive}{17.84}(0\%) \\

  \bottomrule  
\end{tabular}
\end{table*}
 
\section{Conclusion}
%在本文中，我们发现并证明了大数据大参数的分割大模型涌现出来的抗雾能力。我们将这种能力应用在了无法大参数和大训练集去雾小模型上。首先，我们将分割大模型的分割结果灰度编码。然后将这些分割mask输入到去雾小模型新的通道上，从而获得更好的去雾结果。同时，我们还发现了大模型掩膜对于去雾网络训练的加速作用。我们分析了不同去雾场景，不同数据集，不同分割大模型型尺寸和不同去雾小模型尺寸对于去雾结果的影响。对于复杂的雾霾场景，大模型能帮助小模型更好的处理。广泛的实验结果表明我们的方法表现良好。
In this article, we discover and prove the emergence of anti-fog ability in large-dataset, large-parameter segmentation model. We apply this ability to small dehaze models. First, we gray-scale encode the segmentation results of the large-scale segmentation model. Then, we input these segmentation masks into a new channel of the small dehaze model to achieve better dehaze results. At the same time, we also discover the acceleration effect of large model masks on the training of the dehaze network. We analyze the impact of different fog scenarios, dehaze datasets, large-capacity segmentation model sizes, and small dehaze model sizes on dehaze results. For complex fog scenes, large models can help small dehaze models better handle them. A wide range of experimental results show that our method performs well.

%总体来说，我们让语义分割大模型的优势来帮助无法进行大参数大模型训练的去雾小模型。让小模型可以享受大模型的各种发展优势，同时也不需要大模型对小模型的任务进行针对优化。我们开辟了一条大模型帮助底层计算机视觉的新方法，让底层的视觉任务也能收益于大模型的发展。
Overall, we enable the large semantic segmentation models to help small dehaze models that cannot be trained with large parameters and large models. This allows small models to enjoy the various development advantages of large models, while also without requiring large models to be specifically optimized for small-model tasks. We propose a new method for large models to help low-level computer vision, allowing low-level visual tasks to benefit from the development of large models.

\printcredits

\bibliographystyle{cas-model2-names}

\bibliography{cas-refs}

\bio{figs/pic1}
Author biography with author photo.
Author biography. Author biography. Author biography.
 
\endbio
\end{document}